\documentclass{article}

\usepackage[english]{babel}

\usepackage[letterpaper,top=2cm,bottom=2cm,left=3cm,right=3cm,marginparwidth=1.75cm]{geometry}
\usepackage{multirow}
\usepackage{amsmath}
\usepackage{graphicx}
\usepackage{array}
\usepackage{adjustbox}
\usepackage{colortbl}

\usepackage[colorlinks=true, allcolors=blue]{hyperref}

\title{Named Entity Recognition for Address Extraction in Speech-to-Text Transcriptions Using Synthetic Data}
\author{Bibiána Lajčinová \textsuperscript{(1)}, Patrik Valábek \textsuperscript{(1), (3)}, Michal Spišiak \textsuperscript{(2)}}
\date{\footnotesize\textsuperscript{\textbf{(1)}}Slovak National Supercomputing Centre, Bratislava, Slovak Republic\\
 \textsuperscript{\textbf{(2)}}nettle, s.r.o., Bratislava, Slovak Republic\\
 \textsuperscript{\textbf{(3)}}Institute of Information Engineering, Automation, and Mathematics, Slovak University of Technology in Bratislava, Slovak Republic\\
}

\begin{document}
\maketitle

\begin{abstract}
This paper introduces an approach for building a Named Entity Recognition (NER) model built upon a Bidirectional Encoder Representations from Transformers (BERT) architecture, specifically utilizing the SlovakBERT model. This NER model extracts address parts from data acquired from speech-to-text transcriptions. Due to scarcity of real data, a synthetic dataset using GPT API was generated. The importance of mimicking spoken language variability in this artificial data is emphasized. The performance of our NER model, trained solely on synthetic data, is evaluated using small real test dataset.
\end{abstract}

\section{Introduction}

Many businesses spend large amounts of resources for communicating with clients. Usually, the goal is to provide clients with information, but sometimes there is also a need to request specific information from them. 

In addressing this need, there has been a significant effort put into the development of chatbots and voicebots, which on one hand serve the purpose of providing information to clients, but they can also be utilized to contact a client with a request to provide some information.

A specific real-world example is to contact a client, via text or via phone, to update their postal address. The address may have possibly changed over time, so a business needs to update this information in its internal client database.

Nonetheless, when requesting such information through novel channels---like chatbots or voicebots---it is important to verify the validity and format of the address. In such cases, an address information usually comes by a free-form text input or as a speech-to-text transcription. Such inputs may contain substantial noise or variations in the address format. To this end it is necessary to filter out the noise and extract corresponding entities, which constitute the actual address. This process of extracting entities from an input text is known as Named Entity Recognition (NER). In our particular case we deal with the following entities: municipality name, street name, house number, and postal code.
This technical report describes the development and evaluation of a NER system \footnote{Available at \href{https://huggingface.co/nettle-ai/slovakbert-address-ner}{https://huggingface.co/nettle-ai/slovakbert-address-ner}} for extraction of such information.

\section{Problem Description and Our Approach}
This work is a joint effort of Slovak National Competence Center for High-Performance Computing and nettle, s.r.o., which is a Slovak-based start-up focusing on natural language processing, chatbots, and voicebots. Our goal is to develop highly accurate and reliable NER model for address parsing. The model accepts both free text as well as speech-to-text transcribed text. Our NER model constitutes an important building block in real-world customer care systems, which can be employed in various scenarios where address extraction is relevant.

The challenging aspect of this task was to handle data which was present exclusively in Slovak language. This makes our choice of a baseline model very limited.

Currently, there are several publicly available NER models for the Slovak language. These models are based on the general purpose pre-trained model SlovakBERT~\cite{slovakbert}. Unfortunately, all these models support only a few entity types, while the support for entities relevant to address extraction is missing. A straightforward utilization of popular Large Language Models (LLMs) like GPT is not an option in our use cases because of data privacy concerns and time delays caused by calls to these rather time-consuming LLM APIs.

We propose a fine-tuning of SlovakBERT for NER. The NER task in our case is actually a classification task at the token level. We aim at achieving proficiency at address entities recognition with a tiny number of real-world examples available. In Section~\ref{section:data} we describe our dataset as well as a data creation process. The significant lack of available real-world data prompts us to generate synthetic data to cope with data scarcity. In Section~\ref{section:model} we propose SlovakBERT modifications in order to train it for our task. In Section~\ref{section:iterative} we explore iterative improvements in our data generation approach. Finally, we present model performance results in Section~\ref{section:results}.

\subsection{Data}\label{section:data}

The aim of the task is to recognize street names, house numbers, municipality names, and postal codes from the spoken sentences transcribed via speech-to-text. Only 69 instances of real-world collected data were available. Furthermore, all of those instances were highly affected by noise, e.g., natural speech hesitations and speech transcription glitches. Therefore, we use this data exclusively for testing. 
Table~\ref{tab:example-data} shows two examples from the collected dataset.

\begin{table}[h!]
\centering
\begin{tabular}{p{4cm}p{3.5cm}p{3.5cm}}
\textbf{Sentence} & \textbf{Tokenized text} & \textbf{Tags} \\
\hline
\multirow{3}{*}{Stupava Záhumenská 834} & Stupava & B-Municipality \\
& Záhumenská & B-Street \\
& 834 & B-Housenumber\\[1.5ex]
\multirow{4}{*}{Ďalšie bauerová 44 Košice} & Ďalšie & O \\
& bauerová & B-Street \\
& 44 & B-Housenumber\\ 
& Košice & B-Municipality\\ 
\end{tabular}
\caption{Two example instances from our collected real-world dataset. The Sentence column showcases the original address text. The Tokenized text column contains tokenized sentence representation, and the Tags column contains tags for the corresponding tokens. Note here that not every instance necessarily contains all considered entity types. Some instances contain noise, while others have grammar/spelling mistakes: The token ``Ďalšie'' is not a part of an address and the street name ``bauerová'' is not capitalized.}
\label{tab:example-data}
\end{table}

Artificial generation of training dataset occurred as the only, but still viable option to tackle the problem of data shortage.
Inspired by the 69 real instances, we programmatically conducted numerous external API calls to OpenAI to generate similar realistic-looking examples. BIO annotation scheme \cite{bioscheme} was used to label the dataset. This scheme is a method used in NLP to annotate tokens in a sequence as the beginning (B), inside (I), or outside (O) of entities. We are using 9 annotations: \textit{O}, \textit{B-Street}, \textit{I-Street}, \textit{B-Housenumber}, \textit{I-Housenumber}, \textit{B-Municipality}, \textit{I-Municipality}, \textit{B-Postcode}, \textit{I-Postcode}.

We generated data in multiple iterations as described below in Section~\ref{section:iterative}. Our final training dataset consisted of more than $10^4$ sentences/address examples. For data generation we used GPT-3.5-turbo API along with some prompt engineering.
Since the data generation through this API is limited by the number of tokens---both generated as well as prompt tokens---we could not pass the list of all possible Slovak street names and municipality names within the prompt. Hence, data was generated with placeholders \texttt{streetname} and \texttt{municipalityname} only to be subsequently replaced by randomly chosen street and municipality names from the list of street and municipality names, respectively. A complete list of Slovak street and municipality names was obtained from the web pages of the Ministry of Interior of the Slovak republic \cite{adresy}. 

With the use of OpenAI API generative algorithm we were able to achieve organic sentences without the need to manually generate the data, which sped up the process significantly. However, employing this approach did not come without downsides. Many mistakes were present in the generated dataset, mainly wrong annotations occurred and those had to be corrected manually. 
The generated dataset was split, so that 80\% was used for model’s training, 15\% for validation and 5\% as synthetic test data, so that we could compare the performance of the model on real test data as well as on artificial test data.

\subsection{Model Development and Training}\label{section:model}

Two general-purpose pre-trained models were utilized and compared: SlovakBERT~\cite{slovakbert} and a distilled version of this model~\cite{distilbert_hf}. Herein we refer to the distilled version as DistilSlovakBERT. SlovakBERT is an open-source pretrained model on Slovak language using a Masked Language Modeling (MLM) objective. It was trained with a general Slovak web-based corpus, but it can be easily adapted to new domains to solve new tasks~\cite{slovakbert}. DistilSlovakBERT is a pre-trained model obtained from SlovakBERT model by a method called knowledge distillation, which significantly reduces the size of the model while retaining 97\% of its language understanding capabilities.

We modified both models by adding a token classification layer, obtaining in both cases models suitable for NER tasks. The last classification layer consists of 9 neurons corresponding to 9 entity annotations: We have 4 address parts and each is represented by two annotations – beginning and inside of each entity, and one for the absence of any entity. The number of parameters for each model and its components are summarized in Table~\ref{tab:model_params}.

\begin{table}[h!]
    \centering
    \begin{tabular}{lrrr}
    \textbf{Base model} & \textbf{Parameters} & \textbf{Classification head} & \textbf{Total} \\
    \hline
    SlovakBERT & 124,054,272 & 6,921 & 124,061,193\\
    DistilSlovakBERT & 81,527,040 & 6,921 & 81,533,961\\
    \end{tabular}
    \caption{The number of parameters in our two NER models and their respective counts for the base model and the classification head.}
    \label{tab:model_params}
\end{table}

Models’ training was highly susceptible to overfitting. To tackle this and further enhance the training process we used linear learning rate scheduler, weight decay strategies, and some other hyperparameter tuning strategies.

Computing resources of the HPC system Devana, operated by the Computing Centre, Centre of operations of the Slovak Academy of Sciences were leveraged for model training, specifically utilizing a GPU node with 1 NVidia A100 GPU. For a more convenient data analysis and debugging, an interactive environment using OpenOnDemand was employed, which allows researches remote web access to supercomputers.

The training process required only 10-20 epochs to converge for both models. Using the described HPC setting, one epoch's training time was on average $20$ seconds for $9492$ samples in the training dataset for SlovakBERT and $12$ seconds for DistilSlovakBERT. Inference on $69$ samples takes $0.64$ seconds for SlovakBERT and $0.37$ seconds for DistilSlovakBERT, which demonstrates model's efficiency in real-time NLP pipelines.

\subsection{Iterative Improvements}\label{section:iterative}

Although only 69 instances of real data were present, the complexity of it was quite challenging to imitate in generated data. The generated dataset was created using several different prompts, resulting in 11,306 sentences that resembled human-generated content.
The work consisted of a number of iterations. Each iteration can be split into the following steps: generate data, train a model, visualize obtained prediction errors on real and artificial test datasets, and analyze. This way we identified patterns that the model failed to recognize. Based on these insights we generated new data that followed these newly identified patterns. The patterns we devised in various iterations are presented in Table~\ref{tab:iteracie}. 
With each newly expanded dataset both of our models were trained, with SlovakBERT's accuracy always exceeding the one of DistilSlovakBERT's. Therefore, we have decided to further utilize only SlovakBERT as a base model.

\begin{table}[h!]
\centering
\begin{tabular}{cl}
\textbf{Iteration} & \textbf{Prompt} \\\hline
1. & Street + House Number + Municipality + Postal Code (+shuffling and omitting) \\
2. & Municipality + Street + House Number + Postal Code (+omitting)\\
3. & Municipality + House Number + Street + Postal Code (+omitting)\\
4. & Municipality + House Number + Postal Code\\
5. & Street + Municipality + House Number (verbal form) + Postal Code (+shuffling) \\
6. & Municipality + House Number + Postal Code (Municipality mentioned twice; +shuffling) \\
7. & All data duplicated to lowercase
\end{tabular}
\caption{\label{tab:iteracie}The iterative improvements of data generation. Each prompt was used twice: First with and then without noise, i.e.,  natural human speech hesitations. Sometimes, if mentioned, prompt allowed to shuffle or omit some address parts.}
\end{table}

\section{Results}\label{section:results}

The confusion matrix corresponding to the results obtained using model trained in Iteration~1 (see Table~\ref{tab:iteracie})---is displayed in Table~\ref{tab:confustionmatrix1}. This model was able to correctly recognize only 67.51\% of entities in test dataset. Granular examination of errors revealed that training dataset does not represent the real-world sentences well enough and there is high need to generate more and better representative data. In Table~\ref{tab:confustionmatrix1} it is evident, that the most common error was identification of a municipality as a street. We noticed that this occurred when municipality name appeared before the street name in the address. As a result, this led to data generation with Iteration 2 and Iteration 3.

\begin{table}[h!]
\begin{adjustbox}{center}
    \begin{tabular}{cr|ccccccccc}
    \multicolumn{2}{c|}{} & \multicolumn{9}{c}{\textbf{Predicted}} \\
    & & \rotatebox[origin=c]{90}{O} &\rotatebox[origin=c]{90}{B-Street} & \rotatebox[origin=c]{90}{I-Street}& \rotatebox[origin=c]{90}{B-Housenumber}& \rotatebox[origin=c]{90}{I-Housenumber}& \rotatebox[origin=c]{90}{B-Municipality}&\rotatebox[origin=c]{90}{I-Municipality} &\rotatebox[origin=c]{90}{B-Postcode} &\rotatebox[origin=c]{90}{I-Postcode} \\
    \hline
    \multirow{9}{*}{\rotatebox[origin=c]{90}{\textbf{Ground truth}}} 
    & O & \cellcolor[rgb]{0.7,0.9,0.7}53 &\cellcolor[rgb]{0.95,0.7,0.7}6 &\cellcolor[rgb]{0.95,0.7,0.7}10 &\cellcolor[rgb]{0.98,0.85,0.85}1 &\cellcolor[rgb]{0.98,0.85,0.85}1 &\cellcolor[rgb]{0.98,0.85,0.85}2 &0 &0 &0 \\
    & B-Street &\cellcolor[rgb]{0.98,0.85,0.85}1 &\cellcolor[rgb]{0.7,0.9,0.7}30 &\cellcolor[rgb]{0.9,0.5,0.5}21 &0 &0 &0 &0 &0 &0 \\
    & I-Street&0 &\cellcolor[rgb]{0.98,0.85,0.85}1 &\cellcolor[rgb]{0.7,0.9,0.7}10 &0 &0 &0 &0 &0 &0 \\
    & B-Housenumber & \cellcolor[rgb]{0.98,0.85,0.85}2& \cellcolor[rgb]{0.98,0.85,0.85}1&0 &\cellcolor[rgb]{0.7,0.9,0.7}69 &0 &0 &0 &0 &0 \\
    & I-Housenumber&0 &0 &0 &\cellcolor[rgb]{0.98,0.85,0.85}1 &\cellcolor[rgb]{0.7,0.9,0.7}18 &0 &0 &0 &0 \\
    & B-Municipality& \cellcolor[rgb]{0.95,0.7,0.7}6& \cellcolor[rgb]{0.9,0.5,0.5}37&\cellcolor[rgb]{0.98,0.85,0.85}3 &0 &0 &\cellcolor[rgb]{0.7,0.9,0.7}25 &0 &0 &0 \\
    & I-Municipality&\cellcolor[rgb]{0.98,0.85,0.85}1 &0 &\cellcolor[rgb]{0.95,0.7,0.7}9 &0 &0 &0 &\cellcolor[rgb]{0.7,0.9,0.7}8 &0 &0 \\
    & B-Postcode& 0& 0&0 &0 &0 &0 &0 &\cellcolor[rgb]{0.7,0.9,0.7}1 &0 \\
    & I-Postcode& 0& 0& 0& 0& 0&0 &0 &0 &\cellcolor[rgb]{0.7,0.9,0.7}0 \\
    \end{tabular}
    \end{adjustbox}
    \caption{
    Confusion matrix of model trained on dataset from the first iteration, reaching model's predictive accuracy of 67.51\%. 
}
    \label{tab:confustionmatrix1}
\end{table}

This process of detailed analysis of prediction errors and subsequent data generation accounts for most of the improvements in the accuracy of our model. 
The goal was to achieve more than 90\% accuracy on test data. Model's predictive accuracy kept increasing with systematic data generation. Eventually, the whole dataset was duplicated, with the duplicities being in uppercase/lowercase. (The utilized pre-trained model is case sensitive and some test instances contained  street and municipality names in lowercase.) This made the model more robust to the form in which it receives input and led to final accuracy of 93.06\%. Confusion matrix of the final model can be seen in Table~\ref{tab:confustionmatrix2}.

\begin{table}[h!]
\begin{adjustbox}{center}
    \begin{tabular}{cr|ccccccccc}
    \multicolumn{2}{c|}{} & \multicolumn{9}{c}{\textbf{Predicted}} \\  
    & & \rotatebox[origin=c]{90}{O} & \rotatebox[origin=c]{90}{B-Street} & \rotatebox[origin=c]{90}{I-Street}& \rotatebox[origin=c]{90}{B-Housenumber}& \rotatebox[origin=c]{90}{I-Housenumber}& \rotatebox[origin=c]{90}{B-Municipality}&\rotatebox[origin=c]{90}{I-Municipality} &\rotatebox[origin=c]{90}{B-Postcode} &\rotatebox[origin=c]{90}{I-Postcode}\\
    \hline
    \multirow{9}{*}{\rotatebox[origin=c]{90}{\textbf{Ground truth}}} 
    & O &\cellcolor[rgb]{0.7,0.9,0.7}61 & \cellcolor[rgb]{0.98,0.85,0.85}1&\cellcolor[rgb]{0.98,0.85,0.85}1 &0 &0 &\cellcolor[rgb]{0.98,0.85,0.85}5 &\cellcolor[rgb]{0.98,0.85,0.85}4 &\cellcolor[rgb]{0.98,0.85,0.85}1 &0 \\
    & B-Street & 0&\cellcolor[rgb]{0.7,0.9,0.7}50 & 0&0 &0 &\cellcolor[rgb]{0.98,0.85,0.85}1 &\cellcolor[rgb]{0.98,0.85,0.85}1 &0 &0 \\
    & I-Street& 0& 0& \cellcolor[rgb]{0.7,0.9,0.7}10& 0&0 &0 &\cellcolor[rgb]{0.98,0.85,0.85}1 &0 &0 \\
    & B-Housenumber & 0&0 &0 &\cellcolor[rgb]{0.7,0.9,0.7}72 & 0& 0& 0& 0& 0\\
    & I-Housenumber&0 & 0&0 & 0& \cellcolor[rgb]{0.7,0.9,0.7}19& 0& 0& 0& 0\\
    & B-Municipality&\cellcolor[rgb]{0.98,0.85,0.85}1 &\cellcolor[rgb]{0.98,0.85,0.85}3 &0 &0 &0 &\cellcolor[rgb]{0.7,0.9,0.7}66 & \cellcolor[rgb]{0.98,0.85,0.85}1& 0& 0\\
    & I-Municipality& 0& 0& \cellcolor[rgb]{0.98,0.85,0.85}1&0 &0 &\cellcolor[rgb]{0.98,0.85,0.85}1 &\cellcolor[rgb]{0.7,0.9,0.7}16 & 0&0 \\
    & B-Postcode&0 &0 &0&0&0&0&0&\cellcolor[rgb]{0.7,0.9,0.7}1 & 0\\
    & I-Postcode&0 &0 &0 & 0&0 &0 &0 &0 & \cellcolor[rgb]{0.7,0.9,0.7}0\\
    \end{tabular}
    \end{adjustbox}

    \caption{Confusion matrix of the final model with the predictive accuracy of 93.06\%. Comparing the results to the results in Table \ref{tab:confustionmatrix1}, we can see that the accuracy increased by 25.55\%.}
    \label{tab:confustionmatrix2}
\end{table}

There are still some errors; notably, tokens that should have been tagged as \textit{outside} were occasionally misclassified as \textit{municipality}. We have opted not to tackle this issue further, as it happens on words that may resemble subparts of our entity names, but, in reality, do not represent entities themselves. See an example below in Table~\ref{tab:example_wrong_prediction}.

\begin{table}[h!]
    \centering
    \begin{tabular}{p{2.5cm}p{2.8cm}p{3cm}p{3cm}}
    \textbf{Sentence} & \textbf{Tokenized text} & \textbf{Tags} & \textbf{Predicted tags}\\
    \hline
    \multirow{4}{*}{Kalša to Kal sa} & Kalša & \cellcolor[rgb]{0.7,0.9,0.7}B-Municipality & \cellcolor[rgb]{0.7,0.9,0.7}B-Municipality \\
    & to & \cellcolor[rgb]{0.7,0.9,0.7}O & \cellcolor[rgb]{0.7,0.9,0.7}O \\
    & Kal & \cellcolor[rgb]{0.95,0.7,0.7}O & \cellcolor[rgb]{0.95,0.7,0.7}B-Municipality\\
    & sa & \cellcolor[rgb]{0.7,0.9,0.7}O & \cellcolor[rgb]{0.7,0.9,0.7}O \\
    \\[-0.5ex]
    \multirow{3}{*}{Košice Hlavná 7} & Košice & \cellcolor[rgb]{0.7,0.9,0.7}B-Municipality & \cellcolor[rgb]{0.7,0.9,0.7}B-Municipality \\
    & Hlavná & \cellcolor[rgb]{0.7,0.9,0.7}B-Street & \cellcolor[rgb]{0.7,0.9,0.7}B-Street \\
    & 7 & \cellcolor[rgb]{0.7,0.9,0.7}B-Housenumber & \cellcolor[rgb]{0.7,0.9,0.7}B-Housenumber \\
    \end{tabular}
    \caption{Examples of the final model's predictions for two test sentences. The first sentence contains one incorrectly classified token: the third token ``Kal'' with ground truth label \textit{O} was predicted as \textit{B-Municipality}. The misclassification of ``Kal'' as a municipality occurred due to its similarity to subwords found in ``Kalša'', but ground truth labeling was based on context and authors' judgment.
    The second sentence has all its tokens classified correctly.}
    \label{tab:example_wrong_prediction}
\end{table}

\section{Conclusions}
In this technical report we trained a NER model built upon SlovakBERT pre-trained LLM model as the base. The model was trained and validated exclusively on artificially generated dataset.
This well representative and high quality synthetic data was iteratively expanded. 
Together with hyperparameter fine-tuning this iterative approach allowed us to reach predictive accuracy on real dataset exceeding 90\%. Since the real dataset contained a mere 69 instances, we decided to use it only for testing. Despite the limited amount of real data, our model exhibits promising performance. 
This approach emphasizes the potential of using exclusively synthetic dataset, especially in cases where the amount of real data is not sufficient for training.

This model can be utilized in real-world applications within NLP pipelines to extract and verify the correctness of addresses transcribed by speech-to-text mechanisms. In case a larger real-world dataset is available, we recommend to retrain the model and possibly also expand the synthetic dataset with more generated data, as the existing dataset might not represent potentially new occurring data patterns. 

\section{Acknowledgements}
The research results were obtained with the support of the Slovak National competence centre for HPC, the EuroCC 2 project and Slovak National Supercomputing Centre under grant agreement 101101903-EuroCC 2-DIGITAL-EUROHPC-JU-2022-NCC-01.

\bibliographystyle{unsrt}
\bibliography{sample}

\end{document}